\documentclass[11pt,a4paper]{article}
\usepackage[hyperref]{emnlp2020}
\usepackage{times}
\usepackage{latexsym}

\usepackage{amsmath}
\usepackage{graphicx}
\usepackage{booktabs}
\usepackage{multirow}
\usepackage{enumitem}
\usepackage{colortbl}
\usepackage{xcolor}
\usepackage{url}
\usepackage{amssymb}
\usepackage{amsthm}
\usepackage{mathtools}
\usepackage[ruled,vlined]{algorithm2e}
\usepackage{subcaption}
\usepackage{makecell}
\usepackage{xspace}
\usepackage{bbm}

\usepackage{xr}
\makeatletter
\def\thickhline{%
  \noalign{\ifnum0=`}\fi\hrule \@height \thickarrayrulewidth \futurelet
   \reserved@a\@xthickhline}
\def\@xthickhline{\ifx\reserved@a\thickhline
               \vskip\doublerulesep
               \vskip-\thickarrayrulewidth
             \fi
      \ifnum0=`{\fi}}
\makeatother

\newcommand\blfootnote[1]{%
  \begingroup
  \renewcommand\thefootnote{}\footnote{#1}%
  \addtocounter{footnote}{-1}%
  \endgroup
}

\newlength{\thickarrayrulewidth}
\newcommand{\com}[1]{}

\usepackage[normalem]{ulem}
\newcommand\mybar{\kern1pt\rule[-\dp\strutbox]{.8pt}{\baselineskip}\kern1pt}

\newcommand{\dataset}{\textsc{Zest}\xspace}
\newcommand{\finalscore}{12}
\newcommand{\meanperformance}{56}
\newcommand{\humanestimate}{42}

\aclfinalcopy 
\def\aclpaperid{3470} 

\title{Learning from Task Descriptions}

\author{Orion Weller*$^{1}$, Nicholas Lourie$^{2}$,  Matt Gardner$^{2}$,  Matthew E. Peters$^{2}$  \\ 
$^1$Brigham Young University \\
$^2$Allen Institute for Artificial Intelligence \\
\normalsize{\texttt{orionw@byu.edu,\{nicholasl,mattg,matthewp\}@allenai.org}}}

\date{}

\begin{document}
\maketitle

\begin{abstract}
Typically, machine learning systems solve new tasks by training on thousands of examples.
In contrast, humans can solve new tasks by reading some instructions, with perhaps an example or two.
To take a step toward closing this gap, we introduce a framework for developing NLP systems that solve new tasks after reading their descriptions, synthesizing prior work in this area.
We instantiate this framework with a new English language dataset, \dataset, structured for task-oriented evaluation on unseen tasks.
Formulating task descriptions as questions, we ensure each is general enough to apply to many possible inputs, thus comprehensively evaluating a model's ability to solve each task.
Moreover, the dataset's structure tests specific types of systematic generalization.\blfootnote{\textasteriskcentered Work done while at the Allen Institute for AI.}
We find that the state-of-the-art T5 model achieves a score of \finalscore\% on \dataset, leaving a significant challenge for NLP researchers.\footnote{Data, evaluation code, baseline models, and leaderboard at \url{https://allenai.org/data/zest}}
\end{abstract}

\section{Introduction}

The dominant paradigm in supervised NLP today is learning from examples, where machine learning algorithms are trained using a large set of task-specific input-output pairs.
In contrast, humans learn to perform the same task by reading a description, after which they are able to perform the task in a zero-shot manner---indeed, this is how crowd-sourced NLP datasets are constructed.
In this paper, we argue that learning from task descriptions in this way is a necessary attribute of a general purpose NLP system, and we propose it as a new paradigm to train and test NLP systems.

\begin{figure}[h]
    \centering
    \includegraphics[width=0.5\textwidth]{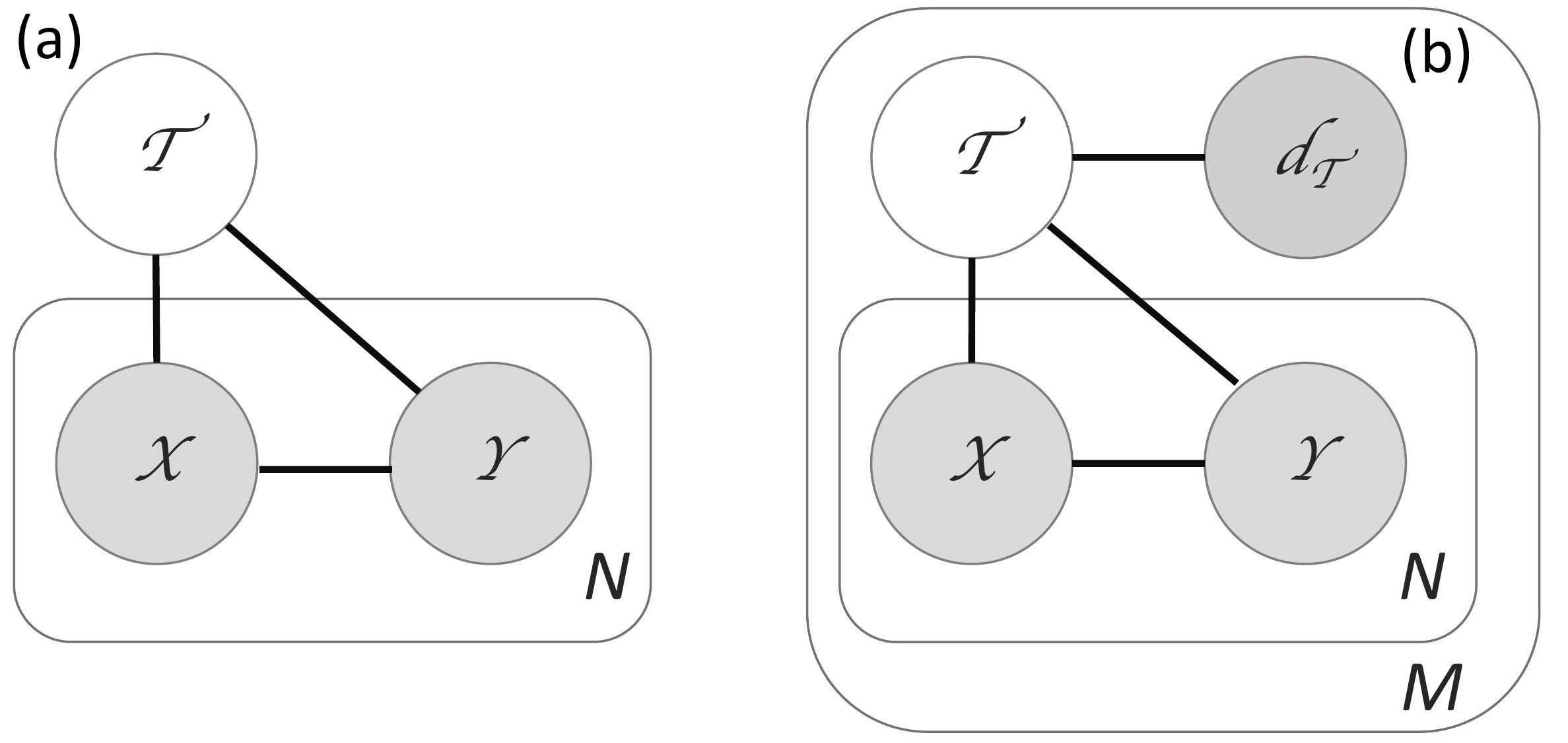}
    \caption{Comparison of (a) supervised learning from examples with observed input $X$, output $Y$, corresponding to an unobserved task $\tau$ (b) our proposed method of learning from task descriptions where systems can make inferences about unseen tasks $\tau$ given a natural language description $d_\tau$.}
    \label{fig:task-formalism}
\end{figure}

Recent work in NLP has shown significant progress in learning tasks from examples. Large pretrained language models have dramatically improved performance on standard benchmarks~\cite{Peters2018,Devlin2018,raffel2019exploring} and have shown promising results in zero shot prediction by leveraging their language understanding capabilities~\cite{levy-etal-2017-zero,zhou-etal-2018-zero,yin-etal-2019-benchmarking}. 

Despite this progress, there are many serious issues that come with learning from examples.  There is an almost infinite number of tasks that a person might wish to solve with a general-purpose NLP system.  Learning to solve these tasks by reading a description instead of observing a collection of examples would solve the problem of having to create training sets for each language task.  Such a system would also be more accessible to practitioners and domain experts in other fields, who could describe their tasks and solve them, opening up new avenues of research where it is expensive or infeasible to gather training data.

Additionally, we find that current supervised learning techniques partly achieve their success due to memorizing uninteresting aspects of the training distribution~\cite{gururangan-etal-2018-annotation,geva-etal-2019-modeling,Gardner2020EvaluatingNM}.  Teaching a system to learn a task from the description alone would alleviate these biases, as new training data would not be needed to learn a novel task.

In this paper, we synthesize prior approaches to zero-shot learning in NLP and provide a formal framework for thinking about the zero-shot prediction problem.  We show that previous zero-shot approaches are limited in both scope of application and rigour of evaluation.  For example, while prior work has used zero-shot prediction for text classification, entity typing, and relation extraction, we push this to the more complex task of slot filling.

We instantiate our formalism in an English language dataset, \dataset (\textbf{ZE}ro \textbf{S}hot learning from \textbf{T}ask descriptions), that is formatted similarly to reading comprehension datasets, in that we formulate task descriptions as questions and pair them with paragraphs of text.  We choose this format as it provides a natural way to crowdsource data.  This zero-shot dataset differs from typical reading comprehension datasets, however, in that each task description is paired with twenty different passages, and we evaluate a model's ability to solve the task, not just give the correct answer for a single (question, passage) pair.  That is, given a question, a model produces some decision function $f$, and it is this function which we comprehensively evaluate on many different inputs.  We also carefully select axes on which to evaluate the generalization of a model to different kinds of task descriptions, changing task descriptions in specific ways to systematically push the field towards more interesting and complex task descriptions.

We evaluate models based on recent state-of-the-art sequence to sequence architectures,
which seem most suited to the task of zero shot prediction in this setting.  We find that our best model based on T5 \cite{raffel2019exploring} achieves a score of only \finalscore\% on this data, leaving a significant gap to our human performance estimate of \humanestimate\%.  Zero shot learning from complex task descriptions remains a significant challenge for current NLP systems.

\section{Learning from task descriptions}
This section describes our framework for enabling zero-shot generalization to unseen tasks, and relates it to prior work.
\label{sec:learning}

\subsection{Learning from examples}
Consider the supervised learning setting\footnote{This setting also includes popular self-supervised objectives such as autoregressive or masked language modeling.} where the goal is to learn a function $y = f_\theta(x)$, with trainable parameters $\theta$, for a particular task.
We define the task $\tau$ as:
\begin{itemize}
    \item a definition for the sets of allowable inputs $x \in \mathcal{X}$, outputs $y \in \mathcal{Y}$, and,
    \item a probability distribution $p_\tau(x, y)$.
\end{itemize}
In text classification, for example, $\mathcal{X}$ is natural language text and $\mathcal{Y}$ is a categorical label from one of $C$ classes.
In the single task setting, the function $f$ is learned by collecting a dataset of labeled examples $\mathcal{D} = \{(x_1, y_1), \ldots (x_N, y_N)\}$ sampled from $p_\tau(x, y)$ (see Fig.~\ref{fig:task-formalism}a).  We call this ``learning from examples''.
Crucially, once $\mathcal{D}$ is constructed, the underlying task definition is discarded, assumed to be captured in the labeled $(x_i, y_i)$ pairs.

There are many ways to sample from $p_\tau(x, y)$ to create a dataset.
One approach, in cases such as language modeling where $p_\tau$ is defined by a set of rules, just applies the rules to raw text.
Another popular approach uses human annotation.
In this case, the most common strategy factorizes $p_\tau(x, y) = p_\tau(y | x) p_\tau(x)$, samples from $p_\tau(x)$ via some method (e.g. collecting text from the domain of interest), and uses a natural language task description, $d_\tau$, to describe $p_\tau(y | x)$.  The description is shown to human annotators who use it to compute $\arg \max_{y \in \mathcal{Y}} p(y | x_0)$ for a given $x_0$.

\subsection{Learning from task descriptions}
The largest downside to learning from examples is that every new task requires collecting a new dataset to learn a new function $f_\theta(x)$ for the task.
This approach also discards the task definition after the labeled dataset is constructed, despite the fact that the task definition carries all of the information necessary for a human to solve the task.
Moreover, it holds the task constant at test time (except in certain limited cases, see Sec.~\ref{sec:framework_discussion}).

Our proposed framework, which we call ``learning from task descriptions'', removes these restrictions.  First, instead of discarding the task definition, we provide a natural language description of it to the model, in addition to the input $x$.
Second, by providing the model with the task description, we expect it to generalize to \textit{unseen tasks} at test time in a zero-shot way.

These modifications shift the learning problem from fitting a probability distribution in the learning from examples approach, to understanding the semantics of a task description in order to apply it to a given input in the learning from task descriptions approach.  Successfully building a model to perform in this manner would open up a wide range of NLP applications whereby one could simply construct an NLP system by describing the desired output in natural language.

Our proposed framework is illustrated in Fig.~\ref{fig:task-formalism}b.
In contrast to learning from examples, we assume the task description $d_\tau$ is observed for $M$ different tasks, and that each of these tasks has some number $N$ of observed $(x_i, y_i)$ pairs.

\subsection{Task competence}
\label{sec:task_competence}
In order to test whether a system can adequately perform an unseen task, we propose a new evaluation metric as follows.
Traditional evaluation metrics in supervised learning are averages over instance-level metrics, that is, they perform independent computation on individual $(x, y)$ pairs and aggregate them across a dataset to produce a summary score.  As we are interested in assessing whether a model can competently perform a task from its description, we instead first evaluate whether a model can perform each individual task using the entire \emph{set} of $(x, y$) pairs for a given task, and then report averages over all tasks.

Formally, a dataset with $M$ tasks can be viewed as the concatenation of $M$ different $N_j$ sized datasets, $\mathcal{D}_j = \{(x_1, y_1), \ldots (x_{N_j}, y_{N_j})\}$, one for each task.
We assume each task has an associated metric $\mu_j(\mathcal{D}_j, f_\theta) \in \mathbb{R}$, which is used to compute the model performance for task $\tau_j$ on $\mathcal{D}_j$ for the model represented by $f_\theta$.  For simplicity, we assume each metric is such that larger values indicate better performance\footnote{This can be achieved by rescaling if necessary.}.
Then, for a given level of competence $c_j$ for task $\tau_j$, we say that the model can perform the task if $\mu_j \ge c_j$.  The final model competence metric is the average individual task competence over the dataset, $c = \frac{1}{M} \sum_j \mathbbm{1}(\mu_j \ge c_j)$, where $\mathbbm{1}$ is the indicator function.  In the special case where $c_j$ has the same threshold $T$ for all $j$, we write ``C@T'' to represent the competence at $T$.

As a concrete example of this metric, consider the simple case where all $M$ tasks are binary classification (so that unseen classes correspond to unseen tasks).  If we adopt accuracy as the metric for all tasks, and set $c_j$ to 90\% for all $j$ then a C@90 of 72\% indicates that the model is able to successfully classify unseen inputs $x$ into a set of unseen classes $\mathcal{Y}$ with at least 90\% accuracy, for 72\% of the unseen tasks $\tau$.

\subsection{Discussion}
\label{sec:framework_discussion}
Prior researchers have recognized the limitations of learning from examples, and have worked to address some of them.  Our proposed framework builds upon and generalizes much of this work.

Zero-shot learning \cite{Chang2008ImportanceOS,socher2013zero,norouzi2013zero} asks systems to generalize to unseen classes at test time.  In this approach, the task is the same at both train and test time---models are only asked to generalize to new classes.  In terms of the graphical model in Fig.~\ref{fig:task-formalism}, prior work attaches a natural language description to some new $y_i$ at test time.  In contrast, our approach asks models to generalize to entire unseen \emph{tasks}, attaching the natural language description to the task variable $\tau$.  Zero-shot learning has been widely adopted including for classification \cite{dauphin2013zero}, entity typing \cite{ma2016label,zhou-etal-2018-zero} and relation extraction \cite{levy-etal-2017-zero,ShiSimpleBERTRELSRL2019}.

More closely related to our approach are the zero-shot experiments in \citet{radford2019language,brown2020language} that provide a generative language model with a prompt (that could be viewed as a type of task description) and asks for a completion.
This is similar to the observation in \citet{Petroni2019LanguageMA} that it is possible to extract knowledge graph relationships from large language models with an appropriate prompt.
\dataset provides a benchmark dataset for systematically measuring how well models can generalize to many tasks in the zero-shot setting.

Multitask learning \cite{caruana1997multitask,collobert2008unified} seeks to learn a single model that can solve multiple tasks simultaneously, similar to our framework that seeks to learn a model that can solve many tasks.  However, in multitask learning each task is learned from examples, and the model is not able to generalize to unseen tasks.
This is also the case for newer control code type approaches \cite{raffel2019exploring,keskarCTRL2019} to multitask learning, where the task is encoded as short string, often containing no information other than a largely meaningless identifier.

There are also connections between our proposed framework and tasks such as natural language inference (NLI) or reading comprehension (RC), where two natural language inputs (a premise and a hypothesis for NLI, and a question and passage for RC) are used to predict some output.  In our case, we have two observed variables, $x$ and $d_\tau$, which influence the prediction of the output $y$ (Fig.~\ref{fig:task-formalism}).  Indeed, the baseline model that we discuss in Section~\ref{sec:model} takes a similar approach to NLI and RC and jointly models the two textual inputs.
This correspondence has been used in prior work, where \citet{yin-etal-2019-benchmarking} used a model pretrained on MNLI~\cite{williams-etal-2018-broad} to perform zero-shot text classification.  A key difference, however, is that hypotheses in NLI and questions in RC are typically only paired with single inputs.  In fact, they typically only make sense for a single input, and thus it is hard to characterize these narrow questions as ``task descriptions''.

Lastly, the problem of learning from task descriptions is fundamentally one of translating a natural language description into some executable function that can operate on arbitrary inputs.  This problem has been well-studied for narrow domains in the semantic parsing literature~\cite{Zelle1996LearningTP,Zettlemoyer2005LearningTM,liang-etal-2011-learning,andreas-etal-2013-semantic}, though the input is typically a single static database, not arbitrary natural language text.  Attempts to generalize semantic parsing to more open domains are still nascent~\cite{Chen2020Neural,Gupta2020Neural}.


\begin{table*}[t!]
\begin{tabular}{p{2.2cm}p{3.5cm}p{5cm}>{\raggedright\arraybackslash}p{3.5cm}}
\toprule
\textbf{Generalization} & \textbf{Question} & \textbf{Input Passage (shortened)} & \textbf{Answer} \\
\midrule
Base & Can I hike to a waterfall at this national park? & ... Yet here at Whiskeytown NRA, we encourage you to chase waterfalls - go visit them! Whiskeytown has four major waterfalls ... & Yes \\
\midrule
Paraphrase & Is there a waterfall to hike to at this national park? & (same as above) & Yes \\
\midrule
Semantic Flips & Can I hike to a canyon at this national park? & ... descending 1,300 feet (396 m) past a large alcove, the trail meanders in a wide canyon ... & Yes \\
\midrule
Composition & What time of year is best to see the popular waterfalls in this national park?  & ... Two viewing platforms provide the best view of Great Falls. This overlook is the last place that the Falls can be viewed ... & NA \\
\midrule
Output Structure & What waterfall hikes are there in this national park and are they wheelchair accessible? & ... Bridalveil Fall is often the first waterfall you'll see when entering ... Although paved, this is trail is not wheelchair accessible due to its grade. & [\{``waterfall hike":``Bridalveil Fall", ``wheelchair accessible": ``No"\}]\\
\bottomrule
\end{tabular}
\caption{Example instances from \dataset.  The composition question is combined with ``What are the popular tourist spots in this national park?" We chose to format the relation extraction questions as JSON, see Section~\ref{sec:modeling} for details.}
\label{tab:generalization}
\end{table*}


\section{Instantiating the Framework}
\label{sec:instantiating_framework}
Section~\ref{sec:learning} showed a framework for training and testing a general purpose system that could perform unseen NLP tasks. An ideal system in this framework would be able to read the descriptions of the tasks in the GLUE suite~\cite{wang2019glue} and perform well with no additional training.  However, this goal is far beyond the current capabilities of today's models. In order to make progress, we must break down the problem into manageable steps.   In this section we outline the scope that we envision for a reasonable NLP-focused dataset that can push forward the current state of learning from task descriptions, without being so challenging as to be out of reach.  Sec.~\ref{sec:zest} describes the data collection process for \dataset, our new English benchmark built following this scope.

To define the scope, we begin by considering the types of applications a model that could successfully learn from task descriptions might enable.  The largest bottleneck in building NLP applications today is collecting labeled data.  Our framework would eliminate this step, making it possible to build ad hoc NLP applications to easily filter, categorize, or extract structured information from corpora.  For example, when planning a camping trip, one might want to know ``What are the names of all the campgrounds and their locations?'' that are listed in a collection of documents, which specifies an ad hoc request to return all examples of the \texttt{located\_at} relationship between the \texttt{campground} and \texttt{location} entity types. 
Accordingly, it's important to include examples of the basic task building blocks of such a system: classification, typed entity extraction, and relation extraction in a benchmark dataset.  In doing so, it would unify the prior work in zero-shot NLP (Sec.~\ref{sec:framework_discussion}) that has focused on just a single task, and require a single model to be able to handle any of these tasks at test time, instead of separate models for each task.

More concretely, as each task $\tau$ defines a set of allowable outputs $y \in \mathcal{Y}$, we can mix multiple output sets $\mathcal{Y}$ in a single dataset as long as the output set is specified in the task description. \dataset includes the most common output sets: discrete classes, lists of (optionally) typed spans from the input, and relationships between spans.
Examples of each are shown in Table \ref{tab:generalization}, where it is clear from the task description which output $\mathcal{Y}$ is expected.
In addition, we also include the \texttt{NA} output \cite{rajpurkar-etal-2018-know}, signifying that it is not possible to solve the task given the input $x$.  For example, if the task asks a model to extract campground names but the input is an unrelated news article, the output is \texttt{NA}.  Being able to correctly identify unsolvable tasks is important in a practical setting where it is not reasonable to expect every possible task to be solvable with every possible input.

To move beyond aggregating existing approaches into a single dataset, recall that in our framework observing the task description $d_\tau$ in addition to the input $x$ allows us to test a model's generalization relative to four variables: $x$, $y$, $\tau$, and $d_\tau$ (Fig.~\ref{fig:task-formalism}).  Motivated by this observation, we propose an approach that systematically varies the task descriptions and inputs while controlling for other sources of variability in order to test whether a system can generalize in multiple ways.  To implement this idea, we begin by collecting a set of task descriptions, $d_\tau$, inputs $x$, and associated outputs, $y$.  This \textit{base} group of instances already allows us to test performance of unseen tasks on unseen input.  We further augment it with four types of controlled generalization: paraphrase, semantic flips, composition, and output structure.  Examples of each type of generalization are given in Table~\ref{tab:generalization}.

\paragraph{Paraphrase}
We can test generalization to changes in the task description $d_\tau$ while keeping the task $\tau$ fixed by paraphrasing the description.  By also fixing $x$, we can use these paraphrases to test whether a model consistently predicts the correct output given the same input and underlying task.  As we collect applicable inputs $x$ for a task using a retrieval mechanism given the task description (Section~\ref{sec:zest}), this also adds some lexical distance between the input and the description, to avoid simple lexical shortcuts to solving the task~\cite{gardner-etal-2019-making}.

\paragraph{Semantic flips}
Closely contrasting examples have long provided an effective means of evaluation in NLP~\cite{levesque2012winograd,sennrich-2017-grammatical}, forcing a model to understand how small changes in inputs correspond to large changes in expected outputs.
We take inspiration from this idea to include task description semantic flips, where a given task is modified in a minimal way (e.g. by changing a single word) to semantically change the meaning of the task. As the description is largely unchanged (including the output set $\mathcal{Y}$), this tests whether systems can distinguish between descriptions that are minimally changed.

\paragraph{Composition}
To further test whether systems can understand a task description, we can compose base tasks into new tasks with operators such as ``and'' and ``or''.  By examining the performance difference between the base group of tasks and the compositionally generated group of tasks we can estimate the extent to which a system can compose tasks in a novel way.

\paragraph{Output structure}
We can also test whether models can generalize to unseen structured outputs $y_1 \in \mathcal{Y}$ where $y_1$ is not seen in the training set.  Among the many ways to accomplish this, we chose a method that asks models to produce output equivalent to slot filling or n-ary relationship extraction in the zero-shot setting.  In this case, task descriptions correspond to a specification of an output structure that includes typed entity and relationship extraction where the entity types and relationships have not been seen in training.


\section{Collecting \dataset}
\label{sec:zest}
To illustrate our novel way of evaluating and framing the ``learning from task descriptions" problem, we provide an empirical demonstration of where current systems fail by collecting a challenge dataset.  We hope this will serve as a starting point for making progress towards this goal of learning from descriptions. In this section we describe our annotation efforts, which consist of our design for the dataset, as well as three crowdsourcing steps: collecting tasks (in question form), gathering relevant documents, and annotating answers for the (task, document) pairs.

\subsection{Dataset Design}
Our dataset consists of base task descriptions which are varied along the four areas of generalization found in Section~\ref{sec:instantiating_framework}, allowing us to systematically control for generalization across the different base tasks. 
We collect annotations for approximately 20 different input documents for each task so that we can calculate the competency metric.

The framework described in Section~\ref{sec:framework_discussion} applies to any task description, thus, it is agnostic to the specific format. In deciding how to format the task descriptions in \dataset we chose to use a question format for the tasks, as crowdsourcing annotations for questions is well established, and a QA format may potentially allow transfer from existing question answering datasets.  We note however, that a declarative task description such as ``return a list of hikes in the national park described in the document" fundamentally asks for the same information as the question ``what are the hikes in this national park?"  As a result, we will use the terms \textit{task description} and \textit{question} interchangeably when discussing our creation of \dataset.

\subsection{Task Generation}
As each question should apply to numerous documents, we used Mechanical Turk\footnote{We initially opened our crowdsourcing pipeline to the U.S. population on Mechanical Turk that had above a 99\% acceptance rate with over 5000 completed HITs, but reduced this pool to only include workers who performed well on initial HITs.} to crowdsource common questions that someone might ask about three different domains: U.S. presidents, dog breeds, and U.S. national parks.  We use multiple domains to include diversity in our tasks, choosing domains that have a multitude of entities to which a single question could be applied. Workers were asked to generate questions that could apply to any entity in that domain and we manually removed questions that contained duplicate meanings to maintain a rich semantic space. This left us with approximately 100 base task descriptions for each domain.  These tasks were generated before gathering input documents, alleviating biases from having workers who had already seen the input passages.

\begin{table}[t]
\begin{tabular}{lrrr}
\toprule
\textbf{Statistic} & \textbf{Train} & \textbf{Dev} & \textbf{Test} \\
\midrule
(task, passage) pairs & 10,766 & 2,280 & 11,980 \\
Avg. passage words & 121 & 122 & 122 \\
Number of tasks & 538 & 114 & 599 \\
Avg. task len [words] & 12.3 & 12.2 & 11.8 \\
\texttt{NA} percent & 0.62 & 0.67 & 0.62 \\
Classification Percent & 0.46 & 0.49 & 0.44 \\
\bottomrule
\end{tabular}
\caption{Summary Statistics for \dataset. Note that \texttt{NA} is the most frequent answer.}
\label{tab:summary_statistics}
\end{table}

We split these tasks into 50\% test, 40\% train, and 10\% development. We then employed other workers to alter them along one of the four areas of generalization. For the paraphrase generation, we asked workers to paraphrase the text so that it retained its original meaning but had a different wording. For the semantic flip questions we asked the workers to keep as much of the task description the same as possible, but to make a slight change that would alter the meaning of the task. Composition tasks were created by randomly sampling three tasks from within each dataset split to combine, letting the worker choose two out of the three.
Tasks for the output structure were created by expanding the base tasks to include multiple structured sub-tasks, using a custom built UI that automatically compiled workers' responses into JSON format.

Each task description created for a particular area of generalization followed its base task to the corresponding dataset split.  Hence the test set contains its own unique base questions as well the derived questions for each area of generalization.

\subsection{Passage Retrieval}
\label{sec:passage_retrieval}
In order to gather a unique set of passages that pertain to a given question, we used Bing and Google Custom Search engines, focusing the results on a narrow subset of webpages.  For U.S. Presidents, our queries were limited to results from Wikipedia pages (for all 45 presidents) as well as information contained on Whitehouse.gov, containing biographies and accomplishments for each President and First Lady.  Similarly, we limited our queries of dog breeds to all 524 pages of Dog Breeds on Wikipedia.  The U.S. National Park passages were retrieved from sub-pages of the National Parks website.  On each of these domains, we ensured that no single entity garnered more than 5\% of the total input documents. Details on how we used these search engines to gather the passages can be found in Appendix~\ref{app:gathering_passages} and in our code.

\subsection{Document Annotations}
We paired the gathered task descriptions with their respective passages and employed our expert workers from Mechanical Turk to annotate the answers.  We had three workers annotate each (task, document) pair.  For the tasks that could be answered with a yes or no response, final answers were chosen by taking the majority answer.  For tasks that involved extracting information from the passage, we used the answer that was the subset of the other answers, preferring shorter responses over longer responses. 
25,026 (task, input, answer) triples, with a total of 1251 task descriptions split across the three domains. These tasks were distributed as 45\% extraction, 45\% classification and 10\% mixed (due to the output structure tasks). More summary statistics can be found in Table~\ref{tab:summary_statistics}.  Our annotation costs were approximately 9,000 USD.


\section{Establishing a Baseline}
\label{sec:model}
This section describes our baseline model results.

\subsection{Evaluation}
\label{sec:evaluation}
Due to class imbalance, we adopt F1 as the metric
when computing the task competency (Sec.~\ref{sec:task_competence}). 
However, to account for partial overlap between model and gold answers, we modify the precision $P$ and recall $R$ as follows. 
Each task $\tau$ has a number of instances $(x_{i}, y_{i})$.  For each instance, we compute a partial overlap score $s_{i}$ that includes an output-type aware\footnote{\dataset includes strings, sets of strings, lists of dicts, and three discrete classes (Yes/No/NA) as valid output types.} best alignment between the model and gold answers and scores individual elements with a word overlap based method.  This is similar to common practice in QA evaluation, extended to handle \dataset's output types.
Then, with \texttt{NA} as the negative class, we compute $P=\sum_i s_{i} / m^+$, $R = \sum_i s_{i} / g^+$ where $m^+$ and $g^+$ are the total model predicted positive (not-\texttt{NA}) and gold positive instances.

\begin{table*}[t!]
\centering
\begin{tabular}{lrrr|rrr}
\toprule
& \multicolumn{3}{c}{\bf Dev} & \multicolumn{3}{c}{\bf Test} \\
& \textbf{Mean} & C@75 & C@90  & \textbf{Mean} & C@75 & C@90\\
\midrule
BART-large \dataset only        &      40 &  13 &  8 &    38 &  11 &  4 \\
T5-11B \dataset only        &      56 &  32 &  12 &    55 &  28 &  11 \\
T5-11B \dataset w/MTL         &    56 &  35 &  14 &    56 &  28 &  12 \\
\midrule     
Human Estimate             &    &      &     & 74 & 61 & 42 \\
\bottomrule
\end{tabular}
\caption{Overall performance of baseline models showing the mean F1 and competency at 75\% and 90\%.  Our best model, a T5 model with multi-task learning from other QA datasets (Section~\ref{sec:modeling}), is only able to perform 12\% of unseen tasks at 90\% F1, compared to a human estimate of 42\% of tasks at 90\% competency.}
\label{tab:results_all}
\end{table*}

\begin{table*}[t!]
\centering
\begin{tabular}{lrrr|rrr}
\toprule
& \multicolumn{3}{c}{\bf Dev} & \multicolumn{3}{c}{\bf Test} \\
\textbf{Generalization Type} & \textbf{Mean} & C@75 & C@90  & \textbf{Mean} & C@75 & C@90\\
\midrule
Base             &    71 &  48 &  16 &    63 &  43 &  22 \\
Paraphrase       &    64 &  36 &  12 &    56 &  32 &  16 \\
Composition      &    66 &  44 &  22 &    65 &  41 &  15 \\
Semantic Flips   &    54 &  27 &   9 &    47 &  18 &   5 \\
Output Structure &    33 &  20 &  10 &    47 &  10 &   3 \\
\midrule
Overall w/MTL          &    56 &  35 &  14 &    56 &  28 &  12 \\
\bottomrule
\end{tabular}
\caption{Detailed T5-11B results for \dataset with multi-task learning using other QA datasets (Section~\ref{sec:modeling}).}
\label{tab:results}
\end{table*}

We take each task's F1 score and evaluate the competency metric for each task, reporting these scores in our final results. Additionally, when tasks are closely related we use a more stringent \emph{consistency} metric~\cite{Gardner2020EvaluatingNM} that computes whether a model is competent in both tasks at the same time.  For paraphrases and semantic flips, our C@T metrics only count a model as competent for a task if it is competent for both the base task description and the changed task description.  This helps to avoid giving the model credit for artificially simple decision boundaries that only accidentally solve the task.

\subsection{Modeling}
\label{sec:modeling}

For baselines, we adopt two recent state-of-the-art models, T5 \cite{raffel2019exploring} and BART \cite{lewis-etal-2020-bart}, both because of their positions on top of popular NLP leaderboards and their text-to-text nature. Beyond training on \dataset{} alone, we also trained T5 using multitask learning (MTL) with a combination of other QA datasets to test transfer to \dataset{}: BoolQ \cite{clark2019boolq}, MultiRC \cite{MultiRC2018}, ReCoRD \cite{zhang2018record}, and SQuAD \cite{rajpurkar2016squad}.

\paragraph{Data Preprocessing} To prepare each task's instances for the model, we prepended ``zeroshot question: '' to the task description and ``zeroshot context: '' to the document, then joined these two parts together with whitespace. For output structure generalization, we formatted the answers as JSON to enable more complex zero-shot relation extraction tasks. Thus, the models output answers as both text and JSON, in a seq-to-seq fashion, depending on the question type. When the question calls for JSON, we deserialize and evaluate it, counting deserialization failures as errors. See Appendix~\ref{app:data_preprocessing} for more on data preprocessing.

\paragraph{Training \& Hyper-parameters}
For T5 11B, our best baseline, training used input and output sequence lengths of 512, a batch size of 32, and grid searched four different learning rates (5e-4, 1e-3, 2e-3, and 4e-3).
See Appendix~\ref{app:training_details} for BART and other T5 details.

\begin{table}[t!]
\centering
\begin{tabular}{lrrrr}
\toprule
\textbf{Input} & \textbf{Mean} & C@75 & C@90 \\
\midrule
Full data     &    56 &  32 &  12 \\
Question only &    12 &  10 &   7 \\
Context only  &     1 &   1 &   1 \\
\bottomrule
\end{tabular}
\caption{T5-11B ablation results on the development set using the full dataset, question only and context only.
Only the overall results are shown.
The context only model predicted \texttt{NA} for each instance.
}
\label{tab:dev_results}
\end{table}

\subsection{Results}

\begin{table*}[t!]
\centering
\begin{tabular}{>{\raggedright}p{1.15cm}>{\raggedright}p{3.5cm}>{\raggedright}p{4.75cm}>{\raggedright}p{2cm}>{\raggedright\arraybackslash}p{2.5cm}}
\toprule
\textbf{Error} & \textbf{Question} & \textbf{Input Passage (shortened)} & \textbf{Predicted} & \textbf{Correct} \\
\midrule
Recall (30\%) & Did this president get a graduate degree?  &  ... at Harvard University, where he earned an M.A. in economics ... & N/A & Yes \\
\midrule
Precision (37\%) & Are the volcanoes in this national park dormant?  & ... Dormant: A volcano that is inactive or resting, but is likely to erupt again in the near future. Extinct: A volcano that has stopped erupting ... & Yes & NA \\
\midrule
Partial (9\%) & What kind of trout can be found at this national park?  & ... The presence of non-native brown trout has the potential to impact brook trout and other native fish populations within several of the park's premier large streams ...  & Brown trout & Brown trout,brook trout \\
\midrule
Other (24\%) & Was this dog breed accepted in the american kennel club in the last twenty years? & ... The Cavalier would go on to be recognized by the American Kennel Club in 1995 ... & No & Yes \\
\bottomrule
\end{tabular}
\caption{
Error distribution of the baseline model. Recall errors are when the model incorrectly predicts N/A; precision errors are when the model should have predicted N/A, but didn't; partial answers are when the model failed to predict all of the members of a list. Other common errors included failing to apply reasoning to answer a question, and predicting the wrong key names when producing JSON outputs.}
\label{tab:errors}
\end{table*}

We present our overall results on \dataset{}, an ablation using T5 to probe for annotation artifacts \citep{gururangan-etal-2018-annotation}, and an error analysis breaking down common mistakes.

\paragraph{Baseline Performance} Table~\ref{tab:results_all} shows the performance of the baselines on \dataset, as well as an estimate of human performance.\footnote{Computed by having an author label answers to 55 tasks from the test set.} We report mean F1 across all \emph{instances} in the data, ignoring their grouping into tasks, as well as our proposed C@T metric, for T $\in \{75, 90\}$. The best T5-11B model has mean performance of \meanperformance\% on the development set, while the BART model has lower scores. Moreover, when we evaluate \emph{task competence}, we see these models only rarely successfully solve the whole task well. For C@90, the T5 model's overall score is only \finalscore\% on the test set.  Multitasking \dataset with other QA datasets only slightly improved results.  Table \ref{tab:results} shows a detailed breakdown of performance across generalization type for the T5 model with multi-tasking.  Detailed results for BART are in the Appendix. Model performance decreases as the generalization difficulty increases from the Base level to Output Structure.  Consistently recovering models from task descriptions alone remains a significant challenge.

\paragraph{Annotation Artifacts \& Ablations} Table~\ref{tab:dev_results} shows ablations on the dev set using T5, illustrating that both the question and context are needed for the model to perform well, as one would expect.  We see that in the context only ablation, the model predicted \texttt{NA} (majority class) for all instances, showing that there were not any systematic biases in the passages alone that the model could exploit.  The context only F1 is non-zero due the fact that one task had all NA answers, which is counted as competent by convention.

\paragraph{Error Analysis} In order to more clearly understand where these models fail, we examined 100 instances of model errors and categorized them. The most frequent errors were when the model failed to recognize the answer (30\% of the time) or predicted something when the answer was \texttt{NA} (37\%).  We provide detailed examples and descriptions in Table~\ref{tab:errors}. Interestingly, the model failed to output parseable JSON on only 1.5\% of all structure questions in the test set and generated a JSON structure format for only 0.008\% of non-structure questions, showing strong results for learning the format for outputting the complex relationships.


\section{Conclusion}
We introduced a framework for creating general purpose NLP systems that can solve tasks from natural language descriptions, synthesizing and extending previous work in zero-shot learning. To make progress toward this goal, we create a dataset, \dataset{}, that rigorously evaluates how well a model truly understands each task. The dataset is designed to test models' ability to systematically generalize across four different areas. State-of-the-art performance on \dataset is \finalscore\%, leaving much room for future improvement.

While we have been focused on zero shot learning from task descriptions, our framework also permits few-shot scenarios where a task description is given along with a handful of examples, making meta-learning approaches applicable.  This is an interesting avenue for future work, for which \dataset{} should also be useful. To facilitate future work, we make our models, code, and data available at \url{https://allenai.org/data/zest}.

\section*{Acknowledgements}
The authors acknowledge helpful feedback from anonymous reviewers and the AllenNLP team. TPU compute used in this work was provided by Google through TensorFlow Research Cloud (TFRC). This research was funded in part by the NSF under awards IIS-1817183 and CNS-1730158.

\bibliography{references, anthology}
\bibliographystyle{acl_natbib}
\appendix

\section{Gathering Passages}
\label{app:gathering_passages}
We used Google and Bing search engines to gather documents for our task descriptions, creating a custom endpoint with a limited number of websites (described in Section \ref{sec:passage_retrieval}) for each domain. Each task description was processed by removing stop words and then used as a query through the respective custom search endpoint. This allowed us to retrieve search snippets and URLs that could be used for further processing.

We used each search snippet to generate the full passage, retrieving the full text of any paragraph that was contained in the snippet, or a random amount (between 0 and 3) of sentences before and after each snippet if the full length of the passage exceeded 300 words.  This ensured that we maintained crucial information from the query while mitigating potential bias from the search engine.

\section{Data Preprocessing}
\label{app:data_preprocessing}
In order to facilitate training a text-to-text model on \dataset{}, we took each task and generated a collection of input-output text pairs. These pairs were then regarded as individual examples in the training---we did not explore approaches that keep examples grouped together by their task. To generate each input, we prepended ``zeroshot question: '' before the task description and ``zeroshot context: '' before the corresponding document. In the case of T5, we appended two newline characters to each and then joined them together, whereas BART used a single space. For each output, we simply used the target \dataset{} provides.

\section{Training Details}
\label{app:training_details}

To facilitate reproducing our experiments, this appendix provides additional details on how we trained and ran predictions for the models.  Code to reproduce the baseline results is available from \url{https://allenai.org/data/zest}.

\subsection{T5 Details} Our baselines build off the T5 11B model \cite{raffel2019exploring}: a text-to-text encoder-decoder structured transformer pretrained via masked language modeling and multi-tasking. T5 11B has 11 billion parameters. Our training, evaluation, and modeling code used the original implementation released with the T5 work.\footnote{\url{https://github.com/google-research/text-to-text-transfer-transformer}}

\paragraph{Training, Evaluation, \& Hyper-parameters} Since T5 frames tasks as text-to-text, the model was trained via teacher forcing \cite{williams1989learning}. Fixed hyper-parameters include an input sequence length of 512, an output sequence length of 512, and a batch size of 32 examples (i.e., instances of tasks, not tasks themselves---see \textbf{Data Preprocessing}). The \dataset{} + \textsc{MTL} baseline equally weighted each component dataset during training, sampling them at the same rate. To tune the learning rate, for each T5 baseline we performed a grid-search over four values: 5e-4, 1e-3, 2e-3, and 4e-3. The best learning rate for each baseline was 4e-3 for context-only, 1e-3 for question-only, 1e-3 for \dataset{}-only (full data), and 5e-4 for \dataset{} + \textsc{MTL}. The model was trained for 25,000 updates with checkpoints taken approximately every 2,500 steps. Throughout training, we kept the 10 most recent checkpoints. All other training specifics were identical to those used in the original T5 work \citep{raffel2019exploring}. For early stopping, we chose the checkpoints with the highest per-instance accuracy on dev to evaluate on test.\footnote{Note that this metric differs from the evaluation we use for reporting results, which is more complex to compute.}

\paragraph{Hardware \& Compute} We trained the T5 models using three v3-256 TPUs on Google Cloud, using one TPU per model and running experiments in parallel. The T5 implementation we built off integrates with Mesh Tensorflow \cite{shazeer2018mesh}, which provides automatic data and model parallelism. For training, we set a model parallelism of 16. All T5 baselines trained the same model (T5 11B), only on different data. Training took 2 hours, 44 minutes, and 38 seconds on average with a standard deviation of 15 minutes and 28 seconds across the 16 runs. Evaluation on the validation set for the \dataset{}-only (full data), context-only, and question-only baselines took on average 26 minutes and 9 seconds with a standard deviation of 1 minute and 4 seconds across 12 runs, while evaluation for the \dataset{} + \textsc{MTL} baseline took on average 45 minutes and 15 seconds with a standard deviation of 1 minute and 11 seconds across 4 runs.\footnote{The multi-task baseline took longer to evaluate because we also evaluated it on the other tasks besides \dataset{}.}

\subsection{BART Details} BART \cite{lewis-etal-2020-bart} is a text-to-text encoder-decoder structured transformer pretrained with a denoising autoencoding objective.  We used the BART-large model with 406 million parameters as implemented in \texttt{transformers}\footnote{\url{https://github.com/huggingface/transformers}}.  We followed the original hyperparameters recommended by the authors for fine tuning for summarization\footnote{\url{https://github.com/pytorch/fairseq/blob/master/examples/bart/README.summarization.md}} with the exception of tuning the batch size, learning rate, number of training epochs, and sequence lengths.  In particular, we used a batch size of 32, maximum source/target sequence lengths of 512/64, four beams for decoding, weight decay of 0.01 and 0.1 label smoothing.  Learning rate was tuned in [3e-5, 5e-5] and number of epochs in [3, 5, 10, 15, 20], with the best model on the development set having learning rate=3e-5 for 15 epochs.  Training took approximately 3.5 minutes per epoch on a single RTX 8000 GPU.  Detailed results for the best model are shown in Table \ref{tab:bart_results}.

\section{Evaluation Details}
\label{app:evaluation-details}

As described in Section~\ref{sec:evaluation}, we evaluate the model in a rigorous manner in order to test how well it truly understands each task. We follow conventions established from previous work in the field \cite{dua2019drop, dasigi2019quoref} in evaluating typical Reading Comprehension benchmarks and expand upon them, to account for novel output structures.

\paragraph{Evaluating Classification}
Classification evaluation is straightforward, taking the modified F1 metric (defined in Section~\ref{sec:evaluation}) of the \texttt{yes}, \texttt{no}, and \texttt{NA} classes.

\paragraph{Evaluating Answer Spans}
We evaluate extracted answer spans by first aligning the gold and predicted answers (in the case of multiple extracted spans) and  computing the F1 word overlap score.  We take the max F1 score (with respect to the different answers given by annotators) as the final score for that prediction. This F1 word-overlap score is calculated from the code of \cite{dua2019drop}.  We then compute the task F1 score following Section~\ref{sec:evaluation}.

\paragraph{Evaluating Output Structure Questions}
As each output structure answer could contain multiple entities (contained in dictionaries, to use JSON terminology.  In Table \ref{tab:generalization} the entity would be ``Bridalveil Fall"), we first align all entities in the predicted and gold answers together. We then use each (key, value) pair as a answer, matching the gold pair to the predicted pair. The score for the value comparisons is evaluated as described in the above two sub-sections, w.r.t whether the value is a classification answer or an extracted answer.  We then weight the value score by the key F1 score, as the key is given in the question and is only a reference to the actual answer (e.g. a model should not receive credit for getting the key right, but should receive a penalty for getting the key wrong).  Each (key, value) pair in all answers for the given task is used in calculating the final task F1 score, as described in Section~\ref{sec:evaluation}.

\section{Baseline Results}
\label{app:baseline-results}

This appendix provides the full results breakdown for T5 trained on \dataset{} alone, BART, and human performance. In addition, we've reproduced Table~\ref{tab:results} in this appendix for easy comparison.

\begin{table*}[t!]
\centering
\begin{tabular}{lrrr|rrr}
\toprule
& \multicolumn{3}{c}{\bf Dev} & \multicolumn{3}{c}{\bf Test} \\
\textbf{Generalization Type} & \textbf{Mean} & C@75 & C@90  & \textbf{Mean} & C@75 & C@90\\
\midrule
Base             &    71 &  48 &  16 &    63 &  43 &  22 \\
Paraphrase       &    64 &  36 &  12 &    56 &  32 &  16 \\
Composition      &    66 &  44 &  22 &    65 &  41 &  15 \\
Semantic Flips   &    54 &  27 &   9 &    47 &  18 &   5 \\
Output Structure &    33 &  20 &  10 &    47 &  10 &   3 \\
\midrule
Overall          &    56 &  35 &  14 &    56 &  28 &  12 \\
\bottomrule
\end{tabular}
\caption{Detailed T5 results for \dataset{} with multi-task training.}
\label{tab:t5_mtl}
\end{table*}

\begin{table*}[b!]
\centering
\begin{tabular}{lrrr|rrr}
\toprule
& \multicolumn{3}{c}{\bf Dev} & \multicolumn{3}{c}{\bf Test} \\
\textbf{Generalization Type} & \textbf{Mean} & C@75 & C@90  & \textbf{Mean} & C@75 & C@90\\
\midrule
Base             &    69 &  48 &  16 &    62 &  40 &  17 \\
Paraphrase       &    56 &  28 &  12 &    56 &  33 &  12 \\
Composition      &    73 &  56 &  22 &    64 &  40 &  15 \\
Semantic Flips   &    56 &  27 &   9 &    45 &  15 &   6 \\
Output Structure &    25 &   0 &   0 &    47 &  14 &   3 \\
\midrule
Overall          &    56 &  32 &  12 &    55 &  28 &  11 \\
\bottomrule
\end{tabular}
\caption{Detailed T5 results for \dataset only training.}
\label{tab:t5_alone}
\end{table*}

\begin{table*}[b!]
\centering
\begin{tabular}{lrrr|rrr}
\toprule
& \multicolumn{3}{c}{\bf Dev} & \multicolumn{3}{c}{\bf Test} \\
\textbf{Generalization Type} & \textbf{Mean} & C@75 & C@90  & \textbf{Mean} & C@75 & C@90\\
\midrule
Base             &    50 &  16 &  8 &    51 &  21 &  7 \\
Paraphrase       &    39 &  8 &  0 &    41 &  13 &  4 \\
Composition      &    44 &  15 &  7 &    44 &  13 &  5 \\
Semantic Flips   &    42 &  5 &   5 &    34 &  7 &   2 \\
Output Structure &    23 &  20 &  20 &    19 &  2 &   2 \\
\midrule
Overall          &    40 &  13 &  8 &    38 &  11 &  4 \\
\bottomrule
\end{tabular}
\caption{Detailed BART-large results for \dataset only training.}
\label{tab:bart_results}
\end{table*}

\begin{table*}[b!]
\centering
\begin{tabular}{lrrr}
\toprule
& \multicolumn{3}{c}{\bf Test} \\
\textbf{Generalization Type} & \textbf{Mean} & C@75 & C@90\\
\midrule
Base             &    88 & 85 & 54 \\
Paraphrase       &    85 & 75 & 50 \\
Composition      &    79 & 67 & 44 \\
Semantic Flips   &    68 & 50 & 40 \\
Output Structure &    51 & 30 & 20 \\
\midrule
Overall          &    74 & 61 & 42  \\
\bottomrule
\end{tabular}
\caption{Detailed human performance on \dataset.}
\label{tab:human_detailed}
\end{table*}

\end{document}